\documentclass{article}

     \usepackage[final]{neurips_2021}

\usepackage[utf8]{inputenc}
\usepackage{amsmath}
\usepackage{graphicx}
\usepackage{booktabs}
\usepackage{rotating}

\usepackage{caption}
\usepackage{subcaption}

\usepackage{lmodern}

\usepackage{tikz}
\usetikzlibrary{positioning}
\definecolor{mycolor}{RGB}{226,185,252}
\definecolor{mycolor2}{RGB}{140,21,21} 
\definecolor{mycolor3}{RGB}{86,1,141} 
\usepackage{mathtools}
\usetikzlibrary{arrows, decorations.markings,shapes,arrows,fit}
\usepgflibrary{shapes.arrows}

\usepackage{hyperref}
\usepackage{authblk}
\usepackage[flushleft]{threeparttable}

\hypersetup{
 colorlinks=true,
 linkcolor=blue,
 filecolor=blue,
 citecolor = black,      
 urlcolor=blue,
 }
     
\usepackage{tikz,tikz-3dplot}
\usetikzlibrary{3d,decorations.text,shapes.arrows,positioning,fit,backgrounds}
\tikzset{pics/fake box/.style args={% #1=color, #2=x dimension, #3=y dimension, #4=z dimension
#1 with dimensions #2 and #3 and #4}{
code={
\draw[gray,ultra thin,fill=#1]  (0,0,0) coordinate(-front-bottom-left) to
++ (0,#3,0) coordinate(-front-top-right) --++
(#2,0,0) coordinate(-front-top-right) --++ (0,-#3,0) 
coordinate(-front-bottom-right) -- cycle;
\draw[gray,ultra thin,fill=#1] (0,#3,0)  --++ 
 (0,0,#4) coordinate(-back-top-left) --++ (#2,0,0) 
 coordinate(-back-top-right) --++ (0,0,-#4)  -- cycle;
\draw[gray,ultra thin,fill=#1!80!black] (#2,0,0) --++ (0,0,#4) coordinate(-back-bottom-right)
--++ (0,#3,0) --++ (0,0,-#4) -- cycle;
\path[gray,decorate,decoration={text effects along path,text={CNN}}] (#2/2,{2+(#3-2)/2},0) -- (#2/2,0,0);
}
}}
\tikzset{circle dotted/.style={dash pattern=on .05mm off 2mm,
                                         line cap=round}}

\newcommand{\GG}[1]{}

\title{Neural Fashion Image Captioning : \\ Accounting for Data Diversity}

\author{%
  Gilles Hacheme \vspace{-0.4cm} \\
  Ai4Innov, Masakhane NLP\\
  \texttt{gilles.hacheme@ai4innov.com}\\
  
  Nour\'eini Sayouti \\
  Ai4Innov \\
  \texttt{noureini.sayouti@ai4innov.com}
  }
  
\usepackage{times}
\usepackage{latexsym}

\usepackage[T1]{fontenc}
\usepackage[utf8]{inputenc}

\usepackage{microtype}

\usepackage{multirow}
\usepackage{rotating}

\usepackage{tikz}
\definecolor{mycolor}{RGB}{226,185,252}
\definecolor{mycolor2}{RGB}{140,21,21} 
\definecolor{mycolor3}{RGB}{86,1,141} 
\usepackage{mathtools}
\usetikzlibrary{arrows, decorations.markings,shapes,arrows,fit, positioning}
\usepgflibrary{shapes.arrows}

\usepackage{lscape}

\usepackage[T4, OT1]{fontenc} % loading the fc package for African characters

\begin{document}
    \maketitle
    \begin{abstract}
    Image captioning has increasingly large domains of application, and fashion is not an exception. Having automatic item descriptions is of great interest for fashion web platforms, sometimes hosting hundreds of thousands of images. 
    This paper is one of the first to tackle image captioning for fashion images. To address dataset diversity issues, we introduced the InFashAIv1 dataset containing almost 16.000 African fashion item images with their titles, prices, and general descriptions. We also used the well-known DeepFashion dataset in addition to InFashAIv1. Captions are generated using the \textit{Show and Tell} model made of CNN encoder and RNN Decoder.
    We showed that jointly training the model on both datasets improves captions quality for African style fashion images, suggesting a transfer learning from Western style data. The InFashAIv1 dataset is released on Github to encourage works with more diversity inclusion. 
    \end{abstract}

    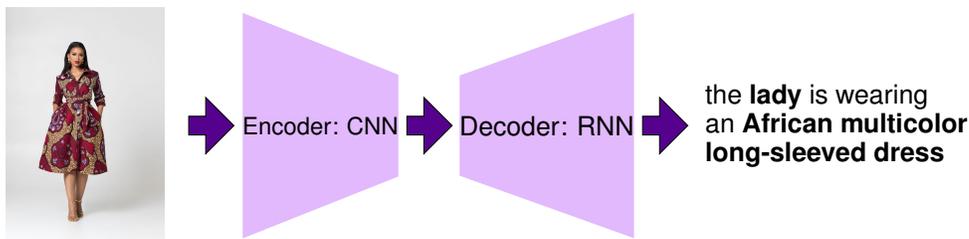
\begin{figure}[ht]
        \tikzset{trapezium stretches=true}
        \begin{tikzpicture}
        	\node[trapezium, fill=mycolor!100, minimum width=30mm , minimum height=15mm,shape border rotate=270,trapezium right angle=86,trapezium left angle=86, anchor = east,font=\sffamily\small,inner sep=0pt] (E) at (0,0)  {Encoder: CNN};
        	
        	\node [trapezium, fill=mycolor!100, minimum width=30mm , minimum height=15mm,shape border rotate=90,trapezium right angle=86,trapezium left angle=86,right=8mm of E, anchor = west,font=\sffamily,inner sep=0pt] (D) at (0,0) { Decoder: RNN};

        	\node[left=1mm of E,single arrow, draw,minimum height=6mm,fill=mycolor3](al) {};
        	\node[align=left, left=1mm of al,font=\sffamily] 
        	{ 
        	{\includegraphics[width=.15\textwidth]{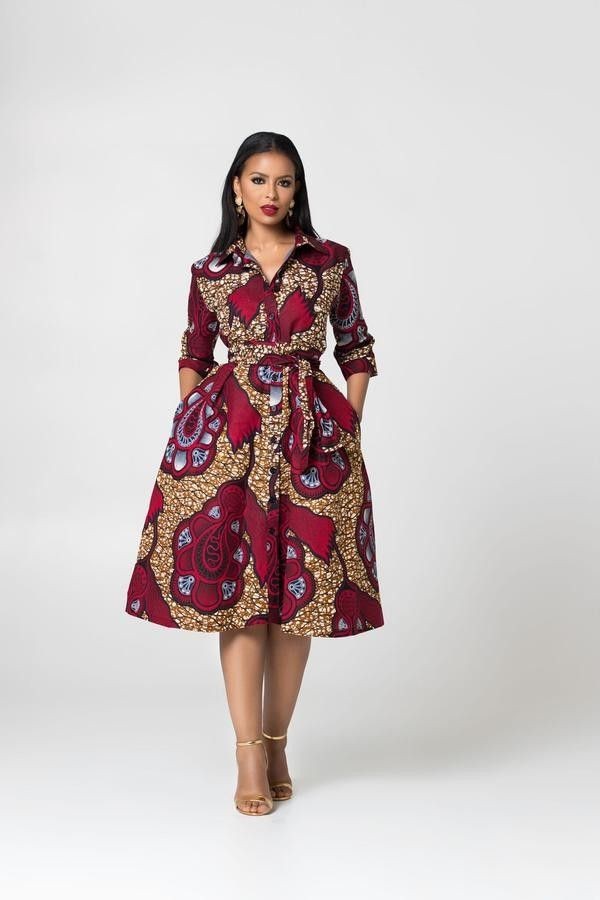}}
    	
        	};
        	\node[right=1mm of E,single arrow, draw,minimum height=6mm,fill=mycolor3](al) {};
        	\node[right=1mm of D,single arrow, draw,minimum height=6mm,fill=mycolor3](ar) {};
        	
        	\node[align=left, right=1mm of ar,font=\sffamily] {the \textbf{lady} is wearing \\ an \textbf{African  multicolor} \\ \textbf{long-sleeved dress}};
        	%%
        	%\foreach \anchor/\placement in
        	%{north west/above left, north/above, north east/above right,
        	%west/left, center/above, east/right,
        	%mid west/right, mid/above, mid east/left,
        	%base west/left, base/below, base east/right,
        	%south west/below left, south/below, south east/below right,
        	%text/left, 10/right, 130/above}
        	%\draw[shift=(E.\anchor),green] plot[mark=x] coordinates{(0,0)}
        	%node[\placement] {\scriptsize\texttt{(E.\anchor)}};
        	%%
        \end{tikzpicture}
        \caption{General structure of the \textit{Show and tell} model}
    \end{figure}

\section{Introduction}
Image captioning is the process aiming to associate a text description to an image in an automatic manner. In this process, a machine is trained to understand the visual content from images and produce corresponding descriptive sentences, the caption. Image captioning is a challenging task, and it recently drew lots of attention from researchers in CV \citep{vinyals2015,Wang2020AnOO,yang2020fashion}. 
Most of the research works on image captioning focus on broad areas with images of different types showcasing humans in their daily life or objects used by them daily \citep{Everingham2009ThePV,Lin2014MicrosoftCC}. However, one of the areas in which image captioning has not been very used is \textit{fashion}, despite all its attractiveness for researchers in AI and specifically in CV. Indeed, in the last decade, the fashion industry attracted many researchers in CV. One reason could be its rapid growth and its vast estimated value.
Indeed, according to \href{https://www.statista.com/}{Statista}, the global apparel market will grow from \$1.5 trillion in 2020 to 2.25 trillion by 2025. Nevertheless, very few papers have applied image captioning to the fashion industry to the best of our knowledge. 
The work by \cite{yang2020fashion} is one of the pioneering works applying image captioning to fashion images. The captions generated by \cite{yang2020fashion} in their paper have both an objective part (description of attributes) and a subjective part aiming to \textit{embellish} outfits descriptions to attract customers' attention and induce them to buy. The usefulness of this type of description is very marketing-oriented. In this paper, we will limit ourselves to generating objective descriptions based on outfits' attributes. Objective description generation is helpful to automate attribute extraction from fashion images. It could be helpful to perform a variety of tasks. Indeed, despite its direct potential usefulness for item description, categorization, and recommendation on fashion websites, it could be used as a building block for more advanced tasks such as image generation from text \citep{zhu2017be}. 
Indeed, for this type of task, objective descriptions are sometimes needed.  Objective descriptions can also be used in any fashion web platform or marketplace to automate image descriptions and classification according to detected attributes. 

In this paper, we are using the model proposed by \cite{vinyals2015} named \textit{Show and tell}, based on the encoder-decoder architecture in which the encoder is a Convolutional Neural Network (CNN) and the decoder is a Recurrent Neural Network (RNN). They initialized the encoder using a pretrained CNN on the ImageNet dataset \citep{deng2009imagenet}, and they used a Long Short-Term Memory (LSTM) as decoder. In this work, we use ResNet152 \citep{he2016deep} as the pretrained model for the encoder.  The encoder-decoder architecture enables us to generate captions containing attributes such as gender, style (here African/Western)\footnote{By African style, we mean any outfit made of African fabric or inspired by traditional African garments.}, color, sleeve type, and garment type.  
The other significant contribution of this paper is that, to the best of our knowledge, unlike most AI research works on fashion, we are including African fashion images in the training set. We worked with a dataset containing both Western and African fashion images. Indeed, it is well known that the generalization capability of AI algorithms is intrinsically linked to the distribution of input data. Low diversified datasets lead to biases, and fashion AI is not an exception. It raises an ethical question as it could result in social exclusion regarding access to AI technologies. It also results in an economic problem, as it limits the adoption of AI technologies by the African fashion industry, for instance. However, the only sub-Saharan fashion market is worth \$31 billion according to the \href{https://www.textiletoday.com.bd/scenario-of-African-fashion-industry/}{Textile Today magazine}. This paper is the first one attempting to address this bias issue, using a dataset containing not only Western but African fashion images. 

Our results showed that captioning quality is higher for Western fashion images due to this data gap. However, we showed that transfer learning occurs as combining both datasets increases African fashion image captioning quality significantly while preserving Western fashion image captioning quality. We exhibited it by comparing joint learning to standalone learning. 

The rest of the paper is structured in five (04) sections. Section 2 explores related work on fashion and ethics in AI. Section 3 introduces InFashAIv1, while Section 5 showcases our results. And finally, section 6 summarizes our main findings and contribution.

\section{Related work}
\textbf{Fashion} has been recently a major topic in Computer Vision (CV)\citep{donati2019fashion, xiao2017fashion, kiapour2014hipster}. \cite{cheng2020fashion} present a survey about more than 200 important studies related to fashion in the field of CV. It covers many topics, including fashion detection, fashion analysis, fashion synthesis, and fashion recommendation. Here we focus on fashion analysis and especially fashion image captioning. Image captioning draws lots of attention from the community. It generally involves attributes prediction. One striking piece of evidence from the literature is that captioning is formalized in some studies as a multi-label classification problem \citep{chen2012describing, yamaguchi2015mix, Chen_2015_CVPR, SUN2016115}. In contrast, others address the problem using an encoder-decoder model where a Convolutional Neural Network (CNN) is used to encode images and a Recurrent Neural Network (RNN), such as Long Short-Term Memory (LSTM), is used to decode descriptions. Descriptions contain desired attributes.  \citep{vinyals2015, xu2015show, herdade2019image, yang2020fashion}
\cite{yang2020fashion} are the first, to our knowledge, applying captioning to fashion images using the encoder-decoder structure. They suggest an improvement of state-of-the-art by introducing Attribute-Level and Sentence-Level Semantic rewards as metrics to enhance generated descriptions relevance. 
They also integrated attribute embedding, Reinforcement Learning (RL) in the description generation process and built the FAshion CAptioning Dataset (FACAD). They achieved state-of-the-art performance compared to previously most known approaches. Nonetheless, their method relies on knowing the attributes of a fashion image to get its description. If it could fit some needs in the fashion industry, it is limited, and cases are more common where only images are available.

In this paper, we used the frame established by \cite{vinyals2015}. \cite{pengshow} provide a robust Pytorch \citep{NEURIPS2019_9015} implementation of \cite{vinyals2015}'s model that we use in this paper. \cite{vinyals2015} showed by their experiment on the MSCOCO 2014 dataset \citep{lin2014microsoft} that the attention mechanism introduced by \cite{xu2015show} enables caption generation that makes more sense but diminishes the generalization ability of the model. Consequently, we did not use the attention mechanism as we are mostly interested in attribute detection. 

\textbf{Ethics} in Artificial Intelligence (AI) technologies is a hot topic today. Ethics in AI can be declined in several aspects, but one of the most critical factors is \textit{inclusiveness} as it directly affects the real-life performance of AI technologies. \cite{chou2018pursuit} suggest \textit{that the most critical step in creating inclusive AI is to recognize where and how bias infects the system}. An important dimension of inclusiveness is the representativeness of datasets used to learn a given task. For example, a face detection algorithm would poorly detect people with a skin color not represented in the training set. 
The same concerns apply to fashion AI technologies. Indeed, fashion AI made significant progress in the last decade, at least partially, thanks to big datasets provided to the community such as Fashion-mnist, Deepfashion, Fashion Landmark Dataset (FLD), Unconstrained Fashion Landmark Database and DeepFashion2 \citep{xiao2017fashion, liu2016deepfashion, liu2016fashion, yan2017unconstrained, ge2019deepfashion2}. But, unfortunately, those datasets contain almost exclusively Western type clothing. This fact limits how they can be used, especially when it comes to processing fashion images of non-Western styles. 
Another concern is the lack of diversity with regards to the skin color of people in the images, the first type of bias raised by \cite{chou2018pursuit}.
One major cause of this lack of diversity in fashion datasets is very likely the gap between the abundance of Western fashion platforms suitable for web scraping and, on the other side, the relative scarcity of similar platforms for African fashion. Still, the lack of diversity could deepen not only social but economic inequalities. Indeed, the African fashion industry could unfortunately not benefit from significant advances offered by AIs trained on under-diversified datasets. 

For all these reasons, we created the Inclusive Fashion AI (InFashAI) project aiming to build more diversified fashion datasets that would be shared with the scientific community.

\section{Towards more diversity in fashion datasets}

The first step of the InFashAI project was gathering data about African fashion. Our major sources of data are \href{https://www.pinterest.com/}{Pinterest} and \href{https://www.afrikrea.com/}{Afrikrea}\footnote{Afrikrea is one of the most known online market places specialized in African fashion, arts and crafts. We have a collaboration with them on the InFashAI project.}. We collected 15,716 item images with their titles, prices (if provided), and general descriptions (if provided). This dataset named \textit{InFashAIv1} is available on Github. 
In this paper, we used a restricted version of InFashAIv1\footnote{We refer further to this restricted version simply as \textit{InFashAI}}. We first only keep images depicting one person. Moreover, to facilitate the learning task, images were further pre-processed in order to remove their backgrounds. We ended up with 8,842 images.
For the purpose of this paper, we generated standardized captions with the following template: the $<$gender$>$ is wearing a/an $<$style$>$ $<$color$>$ $<$sleeve type$>$ $<$garment type$>$.
To generate standardized captions, we needed to annotate images attributes.
We built a small mobile application with \href{https://www.appsheet.com/}{Appsheet}, and we called for volunteers to help us with this task. Our annotated dataset is made of 8,842 images with their standardized captions. In this work, we use a concatenation of this annotated dataset (InFashAI) with the restricted DeepFashion dataset used by \cite{zhu2017be}. This latter dataset is made of 78,979 images with their attributes and captions. We regenerated standardized captions for DeepFashion images using the template presented above. 
We obtained a larger dataset, which should lead to better results on African fashion image captioning.

\section{Results}

We used ResNet152 \citep{he2016deep} as pretrained model for the encoder. 
We tested several sets of parameters to find the one that would allow a \textit{transfer learning} from Western fashion data. As the latter is more abundant, we wanted an architecture leveraging Western style captioning to improve African style captioning. 
Here is the decoder's parameters we finally used: number of layers: 3; hidden size: 512; batch size: 50; dropout: 0.3; learning rate: 0.0002; embedding size: 512; number of epochs: 5. 
We cropped every image to size 224x224. 
To limit noises from InFashAI images backgrounds, we used the tool developed by \href{https://github.com/OPHoperHPO/image-background-remove-tool}{Nikita Selin} to remove them. Most images from DeepFashion were already without any disturbing backgrounds. 
During our experiments, we realized that using a small batch size eases the learning of under-represented features. 
We concatenated African and Western images to form a single pool. This pool contains 87,827 images. Next, we subdivided the shared pool of images into two parts: one for training ( 90 \%  of the dataset, 79,044 images) and the rest for testing (8,783 images).  Both training and test data contain African and Western images.
We subdivided the training set into four ways to obtain different training settings: Train African: the training set only contains African images ( size: 7,802 images); Train Western: the training set only contains Western images ( size: 70,287 images); Train Western + 1/2 African: the training set contains Western images and half of the African images from Train African ( size: 74,188 images); Train All: the training set contains Train Western and Train African ( size: 78,089 images). 
In each training setting, we used a part or the complete training data to train the model. It leads to 4 different versions of the model where the difference only comes from the training data. Hyper-parameters used to train the model are identical for all settings.
We evaluated each model version on the complete test set (Test All: 8,783 images). Results are also reported respectively for African and Western data (Test African and Test Western).

\begin{table}[]
    \centering
    \tiny
    \resizebox{.6\textwidth}{!}{
    \begin{tabular}{lrrr}
    \toprule
    {} &  Test All &  Test African &  Test Western \\
    \midrule
    Train African               &     0.376 &         0.474 &         0.365 \\
    Train Western               &     0.633 &         0.356 &         0.664 \\
    Train Western + 1/2 African &     0.650 &         0.526 &         0.664 \\
    Train All                   &     0.650 &         0.530 &         0.663 \\
    \bottomrule
    \end{tabular}%
    }
    \caption{ BLEU-scores for different configurations}
    %%Note: Train on all: Trained on the concatenation of InFashAI and DeepFashion. 
    \label{tab:BLEU}
\end{table}

Table \ref{tab:BLEU} shows results for the BLEU scores for configurations we considered. When the model is trained on only one style (whether African or Western), it performs unsurprisingly well in terms of BLEU score on the test set of the same style. In other words, when trained on \textbf{Train African} (\textbf{Western}), the model performs better on \textbf{Test African} (\textbf{Western}): 0.474 VS 0.365 (0.356 VS 0.664). 
When we trained the model on \textbf{Train Western + 1/2 African}, we noticed a considerable improvement of the BLEU score on \textbf{Test African} (0.526) while the BLEU score on \textbf{Test Western} remains stable. 
When we trained the model on the whole training set (Train All), results were very closed to the ones obtained with \textbf{Train Western + 1/2 African}. 
These results have profound implications for the amount of data needed to obtain good results on diversified data. Indeed, in our case here, using only half of InFashAIv1 in addition to DeepFashion significantly improves results on African style while preserving results quality for Western style.

\begin{table}[]
    \centering
    \caption{F1 scores for different configurations}

    \resizebox{.45\textwidth}{!}{
    \begin{tabular}{lrrr}
    \toprule
    & \multicolumn{3}{c}{Gender} \\
   {} &  Test All &  Test African &  Test Western \\
   \midrule
   Train African               &     0.816 &         0.911 &         0.817 \\
   Train Western               &     0.972 &         0.848 &         0.986 \\
   Train Western + 1/2 African &     0.978 &         0.910 &         0.986 \\
   Train All                   &     0.976 &         0.912 &         0.984 \\
   \bottomrule
   \end{tabular}%
   }

    \resizebox{.45\textwidth}{!}{
   \begin{tabular}{lrrr}
   \toprule
    & \multicolumn{3}{c}{Style} \\
   {} &  Test All &  Test African &  Test Western \\
   \midrule
   Train African               &     - &         - &         - \\
   Train Western               &     - &         - &         - \\
   Train Western + 1/2 African &     0.997 &         0.994 &         0.999 \\
   Train All                   &     0.997 &         0.994 &         0.999 \\
   \bottomrule
   \end{tabular}%
   }

    \resizebox{.45\textwidth}{!}{
   \begin{tabular}{lrrr}
   \toprule
    & \multicolumn{3}{c}{Color} \\
   {} &  Test All &  Test African &  Test Western \\
   \midrule
   Train African               &     0.140 &         0.072 &         0.148 \\
   Train Western               &     0.661 &         0.229 &         0.717 \\
   Train Western + 1/2 African &     0.678 &         0.287 &         0.725 \\
   Train All                   &     0.685 &         0.296 &         0.732 \\
   \bottomrule
   \end{tabular}%
   }

    \resizebox{.45\textwidth}{!}{
   \begin{tabular}{lrrr}
   \toprule
    & \multicolumn{3}{c}{Sleeve} \\
   {} &  Test All &  Test African &  Test Western \\
   \midrule
   Train African               &     0.312 &         0.492 &         0.293 \\
   Train Western               &     0.841 &         0.635 &         0.864 \\
   Train Western + 1/2 African &     0.839 &         0.695 &         0.855 \\
   Train All                   &     0.823 &         0.689 &         0.838 \\
   \bottomrule
   \end{tabular}%
   }
   
    \resizebox{.45\textwidth}{!}{
   \begin{tabular}{lrrr}
   \toprule
    & \multicolumn{3}{c}{Garment type } \\
   {} &  Test All &  Test African &  Test Western \\
   \midrule
   Train African               &     0.179 &         0.211 &         0.176 \\
   Train Western               &     0.517 &         0.199 &         0.559 \\
   Train Western + 1/2 African &     0.539 &         0.338 &         0.561 \\
   Train All                   &     0.533 &         0.354 &         0.552 \\
   \bottomrule \\
   \end{tabular}%
   }
    \label{tab:f1}
    
    \begin{tablenotes}
        \centering
        \small
        \item Note: Train All: Trained on the concatenation of InFashAI and DeepFashion.
    \end{tablenotes}
    
\end{table}

Nonetheless, the BLEU score is not enough to sufficiently capture the performance of the model. The vocabulary is limited to 69 words, and sentences were generated based on attributes. So, to make a more accurate model assessment in the different settings, we computed F1 scores for each attribute. More precisely, we computed the weighted F1 score as implemented in the Scikit-Learn package to account for the weight of each modality \citep{scikit-learn}. 
Table \ref{tab:f1} shows F1 scores for every configurations we considered. 
The trend observed on the model performance based on the BLUE score also seems to apply to its ability to detect outfits attributes (gender, style, color, sleeve type, and garment type). Therefore, we used the F1 score to evaluate the attribute detection capacity. 
The model trained on data from both clothing styles (Train All) gave quite high F1 scores for color detection when tested on Western clothing (Test Western): 0.732 against 0.717 if the model were only trained on Western data (Train Western). The F1 score obtained on \textbf{Test African} with the \textbf{Train All} model is 0.296 whereas the one obtained with the \textbf{Train African} and \textbf{Train Western} models are respectively 0.072 and 0.229.
The performance gain of the model due to the training data diversity is also observed when detecting sleeve types, particularly for African outfits. However, there is a slight deterioration in the performance of the model in the detection of this attribute when we jump from the \textbf{Train Western + 1/2 African} model to the \textbf{Train All} model. It could partially be due to some noises from errors in African images annotation. 
We also noticed an improvement of the model in the detection of garment types. It is due to the diversity of the training data. Even if we observed a slight deterioration on \textbf{Test Western} when jumping from the \textbf{Train Western + 1/2 African} model to the \textbf{Train All} model, overall, the performance gain due to the training data diversity vastly outweighs this noise effect. 
Additionally, as the model projects images and captions in a common embedding space, we believe that the trained model could generate image embedding and word embedding for tasks involving images and texts. For example, it could be image generation from text or image recommendation from text. Future works will showcase it. 

\section{Conclusion}

This paper is one of the first to tackle neural image captioning in the fashion domain. To build more inclusive fashion AI technologies, we created InFashAIv1, a dataset of almost 16,000 African style fashion images with their titles, prices, and general descriptions. 
We used the \textit{Show and Tell} model with CNN encoder and RNN decoder to generate captions from images. 
Generated captions are satisfying regarding feature detection, especially on Western style images. 
However, the model trained on the augmented dataset showed significant improvement on attributes detection for African images compared to a model trained only on InFashAIv1, suggesting a transfer learning from Western style data (image-caption pairs). 
A significant caveat is that results from our model could be biased towards some stereotypes, like gender stereotypes. Potential biases are not treated in this paper and should be the object of further works. 
Our work is a first step toward building more inclusive fashion AI tools. Though InFashAIv1 is an innovative dataset, its size is somewhat too limited to obtain reliable results and production-ready tools. We are working on extending this dataset to leverage broad research work.
The walk toward inclusive AI has long been stymied by the lack of suitable and large enough datasets. We believe that under-diversified datasets, such as DeepFashion, can be combined with smaller datasets, such as InFashAIv1, to leverage transfer learning and empower less biased and more generalizable models.  InFashAIv1 is released on Github to facilitate future works from the community that, we hope, would integrate more inclusiveness. 

\section*{Acknowledgements}
We are grateful to the \href{https://www.masakhane.io/}{Masakhane} community and particularly to Julia Kreutzer for her constructive comments. We also thank Bonaventure Dossou and Chris Emezue for their helpful feedback. Furthermore, we sincerely thank all the volunteers who helped us to annotate the African images.

\bibliography{biblio}
\bibliographystyle{chicago}

\end{document}